\documentclass[journal,onecolumn]{IEEEtran}
\IEEEoverridecommandlockouts
\usepackage{csquotes}
\usepackage{tabularx,multirow,booktabs,blindtext}
\usepackage{cite}
\usepackage{amsmath,amssymb,amsfonts}
\usepackage{algorithmic,algorithm2e}
\usepackage{graphicx}
\usepackage{textcomp}
\usepackage{xcolor}
\def\BibTeX{{\rm B\kern-.05em{\sc i\kern-.025em b}\kern-.08em
    T\kern-.1667em\lower.7ex\hbox{E}\kern-.125emX}}
\begin{document}

\title{Deep Reinforcement Learning for Detecting Malicious Websites}

\author{\IEEEauthorblockN{Moitrayee Chatterjee and Akbar Siami Namin} \\
\IEEEauthorblockA{Computer Science Department\\
Texas Tech University\\
Lubbock, Texas, USA \\
Email: moitrayee.chatterjee@ttu.edu, akbar.namin@ttu.edu}
}

\maketitle

\begin{abstract}
Phishing is the simplest form of cybercrime with the objective of baiting people into giving away delicate information such as individually recognizable data, banking and credit card details, or even credentials and passwords. This type of simple yet most effective cyber-attack is usually launched through emails, phone calls, or instant messages. The credential or private data stolen are then used to get access to critical records of the victims and can result in extensive fraud and monetary loss. Hence, sending malicious messages to victims is a stepping stone of the phishing procedure. A \textit{phisher} usually setups a deceptive website, where the victims are conned into entering credentials and sensitive information. It is therefore important to detect these types of malicious websites before causing any harmful damages to victims. Inspired by the evolving nature of the phishing websites, this paper introduces a novel approach based on deep reinforcement learning to model and detect malicious URLs. The proposed model is capable of adapting to the dynamic behavior of the phishing websites and thus learn the features associated with phishing website detection. 
\end{abstract}


\section{Introduction}

Phishing is a form of cyber attack typically performed by sending false correspondences that seem to be originated from a legitimate source. The objective of such attack is to gain access to sensitive information such as credit card numbers, credential data, or even to download and activate malware applications and viruses on the target machines. One can say, it is almost essential to have an \textit{online} presence to perform the necessary transactions like banking, e-commerce, social networking. On the other hand, the significance of the World Wide Web has consistently been expanding. The web is not only imperative for individual clients, but also for organizations to function effectively. 

In recent years, the application of various kinds of machine learning algorithms to the classical classification problem and in particular to security and malware detection has received tremendous attention and interest from research community. Furthermore, with the advancement of computational power, deep learning algorithms have created a new chapter in pattern recognition and artificial intelligence. As a result, many classification, decision, and automation problems are now can be formulated through these sophisticated learning algorithms. Deep learning-based approaches are particularly effective when the number of features involved in the computation is large. 

This paper presents a deep reinforcement learning-based model for detecting phishing website by analyzing the given URLs. The model itself is self-adaptive to the changes in the URL structure. The problem of detecting phishing websites is an instance of the classical classification problem. Therefore, we have developed a reinforcement learning model using deep neural network, to solve this classification problem. We have used our model on a balanced and labeled dataset of legitimate and malicious URLs in which 14 lexical features were extracted from the given URLs to train the model. The performance is measured using precision, recall, accuracy and F-measure.
The key contributions of this paper are as follows:
\begin{enumerate}
\item Model the identification of phishing websites through Reinforcement Learning (RL), where an agent learns the value function from the given input URL in order to perform the classification task.
\item Map the sequential decision making process for classification using a deep neural network-based implementation of Reinforcement Learning.
\item Evaluate the performance of the deep reinforcement learning-based phishing URL classifier and compare its performance with the existing phishing URL classifiers.
\end{enumerate}
The proposed approach is robust,  dynamic,  and self-adaptive since reinforcement learning-based algorithms can estimate a solution (i.e., action) based on the stochastic state conversions and the rewards for choosing an action for that state.

The rest of the paper is organized as follows: Section \ref{sec:relatedwork} surveys the related work on phishing identifications using various machine learning based approaches. Section \ref{sec:background} briefly presents the technical background of reinforcement learning and its deep learning variation. Section \ref{sec:model} provides details on the reinforcement learning and the URL structure. Section \ref{sec:experiment} is dedicated to our experimentation details like dataset description and feature extraction and training as well as the results. Section \ref{sec:conclusion} concludes the paper.


\section{Related Work}
\label{sec:relatedwork}

The two most popular phishing detection methods are: (1) {\it Blacklisting} which compares the given URL with the previously reported phishing websites and their URLs status of being malicious or benign. This method is very static meaning that if the URL is a newly created Website then there might be no actual records of it on the Internet, and 2) {\it Analyzing} a given URL based on some heuristics. This technique is a more dynamic method to identify phishing URLs. It parses and extracts features from the URL itself and uses a classifier to decide about a given URL.


Zhang et al.\ \cite{b7} proposed a content-based phishing website detection method called CANTINA. In their proposed framework, the tf-idf score of each term on the web page and generated lexical signatures based on the top five of tf-tdf scores are utilized for deciding about the classification. Then, the lexical signature is provided to a search engine like www.Google.com to look for additional data. If the search query returns the domain name matching the website under consideration, then it is classified as legitimate; otherwise it is classified as a phishing website.  

Xiang et al. \cite{b8} proposed CANTINA+ in which they have used 14 different features categorized in high level webpage features, HTML features, and web-based features. They have applied six different machine learning algorithms on a sample dataset and reported that Bayesian network outperformed the other techniques. As a major drawback, their approach is not resilient to popular attacks such as cross-site scripting attacks.

Both CANTINA and CANTINA+ depend heavily on the text based features and parsing of websites. On the other hand, it is also shown that phishers often construct webpages that contain not only texts but also multimedia data such as flash. As a result, such techniques might be less effective when multimedia components are utilized for the purpose of phishing attacks.   

Abdelhamid et al.\ \cite{b4} proposed a data mining-based approach for phishing URL classification. Their Multi-label Classifier based Associative Classification (MCAC) algorithm functions in three distinct steps: 1) Rules discovery, 2) Classifier building, and 3) Class assignment. In the first step, the algorithm iterates over the training data and uncovers the distinct and salient features. 
In step two, the rules are sorted in order of confidence, length and support to define the classification directive. Finally, in step three, the URLs are classified using the rules with higher support and confidence. 
The authors extracted 16 different features from their sample URLs and tested their algorithm on 1350 websites with 601 legitimate and 752 phishing sites. 

Sahingoz et al. \cite{b2} addressed the phishing URL detection problem using seven different machine learning classification algorithms. Due to the absence of publicly available large dataset of malicious and benign URLs, they prepared a balanced dataset \cite{b3} containing both phishing and benign URLs and made the dataset publicly available. Their work focused on extracting meaningful features from the URLs. 
They extracted NLP-based (Natural language processing) features, word based features and hybrid features during the data pre-processing. Their decision tree-based classifier showed an accuracy of 97.02\% using NLP-based features.

There are some other interesting approaches in detecting phishing attacks and malware using visual similarities \cite{b33, b34}.
\section{Deep Reinforcement Learning: Background}
\label{sec:background}

This section provides a brief overview of the technical aspects of reinforcement learning and its \textit{deep} version. This technique has several interesting applications in different domains \cite{b19, b20}.

\subsection{Reinforcement Learning Paradigm}
The reinforcement learning approach has been utilized to gain proficiency for optimal behavior. This adaptive learning paradigm is defined as the problem of an ``\textit{agent}'' to perform an action based on a ``\textit{trial and error}'' basis through communications with an unknown ``\textit{environment}'' which provides feedback in the form of numerical ``\textit{rewards}'' \cite{b11}. A vanilla form of  reinforcement learning model consists of:

\begin{enumerate}
\item {\it Agent}. An agent learns the model state $S_t$ by reading the input $X_t$, where $t$ denotes the state transitions at time $t$. In the proposed model, the input to the agent  will be the feature vector representation of a given URL. The agent interacts with the learning framework through activities $U_t$ and it gives rewards $R_{(t+1)}$, which can be utilized to improve the \textit{policy} ($\pi$). The reward from these activities is processed and the Q-table is refreshed. Q-table (Q stands for quality) is a reference table or matrix that stores the q-values for a state, action pairs. It is initialized to all zeroes and after each episode, of the learning process, it is updated as the agent learns to take the best action for a state.  
\item {\it Action ($U$).} The actions influence the updates in the environment. The number of activities change based on the feature vectors or the dataset or the number of layers in the neural network. 
\item {\it State ($S$).} At each time step $t$ the state of the environment, the agent is interacting with, changes and affects the action taken by the agent. In this model, a state $s_t$ is determined by the input URL vector $x_t$.
\item {\it Policy ($\pi$).} The policy $\pi$ describes the mapping between the state of the environment and the optimum action (an action pertaining to that state that maximizes the reward) to be performed for that state. The policy set is critical to the the agent of the reinforcement algorithm, as it defines the optimum decision to make.
\item {\it Reward ($R$).} The reward  describes the immediate feedback from the environment,  for an agent, for making the optimum action choice for that particular state. 
\item {\it Discount factor ($\gamma$).}It is defined to balance the performance of the agent, in a way, so that agent can make  optimum choice of actions for both short term and long term rewards. The value of $\gamma$ ranges between $0$ to $1$.
\item {\it Probability of State Transition ($Pr$).} It is the conditional probability ($Pr(s_{t+1}|s_t,u_t)$) for transitioning from state $s_t$ to state $s_{t+1}$. 
\item {\it Episodes.} The number of rounds the agent needs to find the best possible Q-values for all the state, action pairs. 
\end{enumerate}

\subsection{Deep Reinforcement Learning-Based Classifier}
We train a deep neural network as a reinforcement learning agent $A$ that interacts with the environment, receives a training sample $s$ and according to the policy $\pi$ returns the probability of class labels, i.e., action $a$. The policy $\pi$ can be defined as:
\begin{equation}
\pi (a|s)= Pr(a_t = a|s_t = s)
\end{equation}
The goal of the agent is to \textit{explore} and \textit{exploit} the training samples to predict the class labels so as to maximize the cumulative rewards ($R_c$)  through gaining positive rewards as:
\begin{equation}
R_c = \sum_{k=1}^{\infty} \gamma^{k}.r_{t+k} 
\end{equation}
Where $\gamma: \gamma \in \{0,1\}$ is the discount factor. $r$ is the immediate reward and $k$ is the number of episodes.
The Q-value of the state-action $(s,a)$ combination, called the Q-function ,is assessed by applying expected ($E$) reward for following $\pi$: 
\begin{equation} \label{eq:3}
Q^{\pi}(s,a) = E_{\pi}[R_c|(s_t=s, a_t=a)]
\end{equation}
The RL agent  can optimize $R_c$ by solving the optimum $Q^{*}$ function using the $\epsilon$-greedy policy. $Q^{*}$ is the optimal function for optimal policy $\pi ^{*}$. $\epsilon$-greedy chooses a random action uniformly from a set of available actions. The $\epsilon$-greedy approach is used to enable the agent to learn and earn reward from the environment based on the policy $\pi$ so that $Q^{*}$ would be the optimal classifier model for the experiment. The optimal policy $\pi ^{*}$ can be defined as:
\begin{equation}
\pi ^{*} = \bigg\{ \begin{matrix}
1  & a = argmax_a Q^{*}(s,a) \\ 
0 & otherwise  
\end{matrix}
\end{equation}
The Q-function returns the value for taking an action (or predicting a label) for a state (a particular URL vector) under the policy $\pi$. This Q-value (or quality value) is the highest cumulative reward. 
When the actions are limited and state space is small, the Q functions are stored in a table, which would be used to predict the label of a class. However, for higher dimensional data where the state-space combination is too large to record in Q-table, a deep learning network is helpful in learning the optimal classification through gradient decent (policy $\pi$).  A deep learning implementation to approximate the Q values is termed as Deep Q Network (DQN). The DQN uses experience replay for learning. The experience replay is the information about the state transition, action, reward for q-value learning. The learning process uses an \textit{experience memory} (M) to store the information  ($s_1$,$a_t$,$r_t$,$s_{t+1}$) and samples mini batch (Bm) from $M$ to perform gradient decent as per the loss function $L(\theta)$:
\begin{equation}
    L(\theta) = \sum_{(s_1,a_t,r_t,s_{t+1}) \in Bm} (y - Q(s,a,\theta_k))^2
    \label{fig:eqp}
\end{equation}
Where $y$ would be the desired approximation of q-function and takes the form of:
\begin{equation}
    y = \bigg\{ \begin{matrix}
r_j,  & terminal_j = T  \\
r_j + \gamma max_{a_{t_1}}Q(s_{t+1},a_{t+1},\theta_{k-1})), &  terminal_j = F 
\end{matrix}
\label{fig:eqn}
\end{equation}
Where $j$ is a sample from the $M$, \textit{terminal} is the condition when state-action pairs have maximum cumulative rewards and F and T are Boolean values.

\section{A Deep Reinforcement Learning Model}
\label{sec:model}

This section describes the key principles of our proposed algorithm in classifying a given URL as phishing or benign.

\subsection{Problem Statement}
We can formulate the problem of detecting phishing URLs as a binary classification problem, in which the prediction classes are \enquote{phishing} or \enquote{benign}. Let us denote a training dataset with \textit{T} URLs along with data and class labels in the form of ${(u_1,x_1),(u_2,x_2),\ldots(u_T,x_T)}$ where:

\begin{itemize}
\renewcommand\labelitemi{--}
    \item $u_i$ for $i = 1, 2, \ldots T$ denotes a given URL in the training set $T$, and
    \item $x_i \in \{0,1\}$ for $i = 1, 2, \ldots T$ corresponds to the label of the underlying URL where $x_i = 0$ implies benign and $x_i =1$ indicates a phishing URL, respectively.
\end{itemize}

To automate the problem of classification of phishing URLs using deep reinforcement learning, we employ a two-step procedure:
\begin{enumerate}
\item \textbf{Feature Extraction.} The representation of the given URL $u_i$ into a $d$-dimensional (in our problem $d = 14$ vector space of features $v = \{v_1, v_2, \dots, v_i\}$, such that $v_i \in \mathbb{R}^{d}$.
\item \textbf{Deep Reinforcement Learning.} A learning algorithm with a function $f: \mathbb{R}^{d} \to \mathbb{R}$ to predict the class assignment using $v$.
\end{enumerate}

Once the given URL is transformed to its vector representation $v$, the optimization function $f: \mathbb{R}^{d} \to \mathbb{R}$, incorporated into the deep learning part of the algorithm, is applied on $v$ to predict the class label.

\subsection{URL Structure}
\begin{figure}[!t]
  \includegraphics[width=\linewidth]{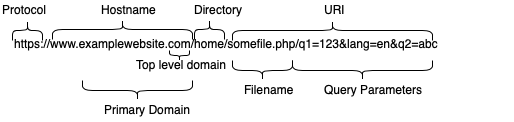}
  \caption{URL structure.}
  \label{fig:url}
\end{figure}

A typical URL has two principle parts: (1) Protocol: Specifies the protocol to be used for communication  between user and web server, (2) Resource identifier: indicating the IP address or the domain space where the resource is located. A colon and two forward slashes separate the protocol from resource identifier, as shown in Figure \ref{fig:url}.

\subsection{Feature Extraction}
There are a certain characteristics of websites that helps in distinguishing between phishing sites from the legitimate ones. Examples of such characteristics include: long URLs, IP address in URLs, and request access to additional URLs in which these characteristics are the indications of being phishing websites. In their work, Mohammad et al. \cite{b5} labeled the website features in four groups: (1) Anomaly-based, (2) Address bar-based, (3) HTML and JavaScript-based and (4) Domain-based. We followed the proposed work in \cite{b5} to build our set of 14 features, as listed below: 

\begin{enumerate}
\item \textit{HTTPS Protocol.} Sensitive information is transferred using HTTPS protocols and utilization of such secure protocol is a typically an indication of being safe. However, phishers also can construct a URL with fake HTTPS protocol. It is necessary to verify if the URL protocol is offered by trusted issuer like VeriSign\footnote{https://www.verisign.com} and thus set this feature to zero. Otherwise, it is set to $1$.
\item \textit{IP Address.} Presence of IP address in the given URL is almost a confirmed indication of the website being a suspicious website (e.g., \textit{ http://149.56.144.216/processa.php}). It is an indicator that the website is trying to gain some unauthorized access. IP addresses are no longer a standard practice for hostnames. Sometimes the IP addresses are converted into hex format for obfuscation. Hence, this feature value is set to $1$ if there exists an IP address in the given URL, or to $0$ otherwise.

\item \textit{Long URLs.} The aim is to construct a URL to be long in order to obfuscate the malicious part. Hence, it is suggested to set this feature to $1$ if the URL is longer than 54 characters and thus classify the given URL as being suspicious. 

\item \textit{URL Containing the @ Symbol.} A browser is designed to ignore everything prior to an @ symbol in a URL. Hence, phishers can redirect a victim to a phishing website using this method. As a result, if a URL contains @ this feature receives the value of $1$.
\item \textit{Adding Prefix or Suffix.} For bypassing the search engine optimization component, phishers often add ``$-$'' to the domain name. Popular search engines such as www.Google.com use ``$-$'' as a word separator. We set this feature to $1$ when there is a ``$-$'' in the domain name.
\item \textit{Sub-domains.} Phishers often add valid sub domain names in the URL to make it appear as a legitimate URL. Hence, check if the number of dots (i.e., ``$.$'') in the hostname is fewer than three and thus set this feature to zero. Otherwise, set it to $1$.
\item \textit{Anchor URLs.} According to \cite{b5} if the webpage has anchors more than 20\% then this feature should be set to $1$ as an indicator of being a phishing website.
\item \textit{Link Hiding.} Phishers obfuscate the actual URL using a fake one on the address bar. This can be identified by MouseOver event. If the MouseOver shows a different URL than the one appear on address bar, this feature should be set to $1$.
\item \textit{DNS Record.} A phishing website generally does not have DNS records. DNS records contain information about the active domain names. Phishing websites are short lived and may not have any DNS record.
\item \textit{Page Redirects.} Phishers redirect the user to another link where the victim could expose sensitive information to them. So, if the number of redirects is greater than $1$ then this feature is an indication of malicious behavior. 
\item \textit{Pop-up Windows.} Legitimate websites do not make their users provide any login credentials through pop-up windows. If the webpage opens more than two pop-up windows, set this feature as $1$ and thus label the webpage as suspicious.
\item \textit{Domain Age.} Newly created websites have high risk of being a phishing website. Using the WHOIS \cite{b10} database, any website younger than one year old should receive a value of $1$ for this feature.
\item \textit{Server from Handlers.} If the webpage asks for user information and redirects the submitted form to a different domain than the one hosting the webpage, it exhibits phishing behavior.
\item \textit{Unusual URLs.} If the URL does not exists in the registered domain names in WHOIS \cite{b10} database then this feature should be set to $1$ to tag the URL as a possible phishing website.
\end{enumerate}
\subsection{Normalization}
All the feature vectors are normalized to binary values that is $0$ and $1$. As well as the class assignment of each URL in training set also $0$ and $1$.
\subsection{Deep Reinforcement Learning-Based Classification}

Reinforcement learning is a robust algorithm that allows an agent to interact with an environment and obtain the states and take action based on that. The proposed classification task is designed as a sequential decision-making problem. Agent for the target classes would receive the input vector representation of the URL. The reward function is not dependent on the class, but both on state and action. When the agent for phishing class would select an action to maximize the reward, the agent from the benign class would try to minimize the reward.

During learning phase, the agent receives the input URL vectors, one at a time and perform the actions (i.e., probabilities of current states to identify the next state) and obtains rewards. At every episodes, the agent learns to obtain more reward. Once the epochs are completed, the model is ready to be applied on unseen data. 

The rewards are based on the actions. For our phishing URL classification problem, the state $s_t$ at time step $t$ is defined by the vector space representation of training dataset $T$. In our case case, it is a $14 * U_T$ matrix, where 14 is the number of feature vectors and $U_T$ is the number of URLs in the dataset. In each episode, the training sample changes.

The action $A = \{0,1\}$ is a binary output referring to the class label (i.e., Phishing or Benign) of the URL $u_t$. If the Q-function value is $\geqslant 0.5$ then we normalize it to 1. In a similar manner, if the Q-function value is between $0$ to $< 0.5$ we normalize it to 0. The reward $R$ is the feedback to the agent from the environment. A reward $r_t$ corresponding to an action $a_t$ signifies if the agent has classified $u_t$ correctly or incorrectly. If the class label of a particular $u_t$ is denoted by $l_t$, then for our experiment, $R$ is defined as: 

\begin{equation}
R = \bigg\{ \begin{matrix}
1,  & a_t = l_t  \\
-1, &  a_t \neq l_t  
\end{matrix}
\end{equation}

\subsection{Training The Network}
The proposed reinforcement-based learning model is generalized to capture the uniformities of the training URL vectors so that the learning agent can earn maximized reward. The agent retains the probabilistic value of successfully predicting the class of each test data. This probability value is used by the agent to learn about the environment. The linear combination of feature vectors is applied to the approximation function for the agent to learn or earn from the environment. The test data set is used continually, so that the agent can form necessary statistics for prediction (or the training algorithm converges), which means there is no new knowledge about the environment that can be learned by the agent to improve prediction. The model is implemented to disregard the issues involving function approximation by a single binary classification for each training data. 

\begin{figure}[!t]
  \includegraphics[width=\linewidth]{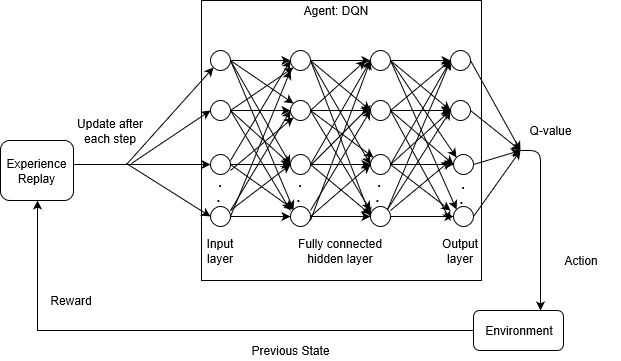}
  \caption{Deep Q Network.}
  \label{fig:dqn}
\end{figure}

We have used deep neural networks to learn from the vector space representation of the phishing URLs. We have one embedding layer and two fully connected layers with ReLU activation and a softmax output layer to implement the DQN or agent of the reinforcement learning based classifier. The controller generates hyper-parameters and the gradient decent leads to updating in policy parameter. The learning rate was $0.001$. A 2-fold cross validation was performed for accuracy improvement. The network architecture is shown in Figure \
\ref{fig:dqn}.

The neural networks are proficient to learn from both linear and nonlinear data. The training and classification algorithm is presented in Algorithm \ref{alg:trn}. This algorithm is based on the original DQN learning algorithm proposed by Mnih et al \cite{b23}. The training agent uses an $\epsilon$-greedy method  to choose an action and gain rewards from the environment. More precisely, the algorithm \textit{greedily} chooses an action with probability $\epsilon$ at random using Equation \ref{eq:3} to reach the action that maximizes the cumulative reward over the epochs of the learning phase.

\begin{algorithm}
\SetAlgoLined
\KwIn{The training samples and their class labels}
\KwOut{The optimum Q-values}
 Initialize experience memory $M$ \;
 Initialize Q-value function with random weights $\theta$\;
  Define episodes $K$\;
  Initialize target Q-value function with random weights $\theta ^{*} = \theta$\;
 \While{episode $1$ to $K$}{
  Initialize $s_1 = v_1$\;
  Initialize the pre-processed sequence function $\phi_1 = \phi(s_1)$ \;
  \While{$t = 1$ to $T$}{
  Perform action $(a_t)$ based on $\epsilon$-greedy, \\ 
  \[
  a_t =  \begin{cases}
  \text{Random-action} &  \text{probability} \epsilon \\ 
  argmax_a Q(\epsilon(s_t),a;\theta) & \text{otherwise} 
\end{cases}
\]
  Observe $r_t$ for action $a_t$\;
  Set $s_{t+1}$ = $s_t$, $a_t$, $v_{t+1}$ \;
  Set pre-processed sequence function $\phi_{t+1}$ = $\phi(s_{t+1})$ \;
  Save ($\phi_1$,$a_t$,$r_t$,$\phi_{t+1}$, $terminal_{j}$) in M\;
  Randomly sample from M and set $y_j$ using equation.\ref{fig:eqn}\;
  Perform gradient decent using equation.\ref{fig:eqp}\;
  \If{$terminal_t = True$}
  {break\;}
  }
 }
 \caption{Training and classification algorithm.}
 \label{alg:trn}
\end{algorithm}

\section{Experimentation Details and Results}
\label{sec:experiment}

The experiment environment for this work was setup on Amazon AWS using EC2 computer instance with an Ubuntu Server 18.04 LTS with Variable ECUs, 1 vCPUs, 2.5 GHz, Intel Xeon Family, 1 GB memory. We installed TensorFlow for implementing the deep reinforcement learning based classifier.

\subsection{Dataset}
The experiment for this work was performed on the Ebbu2017 Phishing Dataset \cite{b3}. The dataset was prepared and publicly made available by Sahingoz et al.  in their phishing URL detection work \cite{b2}. Due to the absence of publicly available large phishing datasets, they prepared a balanced dataset containing both phishing and valid URLs. They have developed their own script to query the internet Yandex Search engine\footnote{https://tech.yandex.com/xml/ } to collect the web pages with higher page rank, and the phishing URLs from  PhishTank\footnote{http://www.phishtank.com }. The dataset contains 73,575 URLs, out of which 36,400 are legitimate and 37,175 are phishing. The training dataset is discrete and contains deterministic classes. The proposed model can make binary predictions of observations of the test data.

\subsection{Evaluation Metrics}
The performance of the proposed model has been assessed using relevance measures like precision, recall, accuracy and F-measure. To calculate these measures, we need to calculate the True Positive (TP), True Negative (TN), False Positive (FP) and False Negative (FN) predictions. TP and TN refer to the correctly classified results; Whereas, FP and FN are the misclassified data. Using these 4 values, we can calculate the performance measures as follows:
\begin{itemize}
\item $Precision = \frac{TP}{TP+FP}$ \\
In classification problem a precision value closer to 1 implies the predicted labels are closer to truth.
\item $Recall = \frac{TP}{TP+FN}$ \\
A recall value closer to 1 implies the all the testing samples could be predicted using the specified model.
\item $Accuracy = \frac{TP+TN}{TP+TN+FP+FN}$\\
A accuracy score closer to 1 implies a high performance of the system.
\item $F-Score = 2*{\frac{Precision * Recall}{Precision + Recall}}$ \\
It is the harmonic mean of precision and recall to signify the model\'s resilience.
\end{itemize}
Table. \ref{tab:relM} shows the average of the relevance measures from different run of the model:
\begin{table}[h]
\caption{Relevance measures of the proposed model.}
\label{tab:relM}
\begin{center}
\begin{tabular}{| c | c | c | c |}
\hline
Precision & Recall & Accuracy & F-Measure \\
\hline
0.867 & 0.88 & 0.901 & 0.873\\
\hline
\end{tabular}
\end{center}
\end{table}
This experiment for the work presented in this paper was performed on a balanced dataset containing both phishing and benign URLs. However, we have not experimented with reward function to explore the performance of the model. The adam optimizer was used for optimizing the parameters of the neural network. We split the dataset into $8:2$ for training and testing purposes.

\section{Conclusion and Future Work}
\label{sec:conclusion}

This paper introduces a reinforcement learning based framework for automated URL- based phishing detection. This deep learning implementation of RL algorithm is a complimentary approach to the existing phishing detection methodologies to make the system dynamic. This work establishes the foundation for a more efficient, dynamic and self-adaptive phishing identification framework. However, this work is not optimized for real world implementation and our future work involves performance tuning for the deep Q learning algorithm to optimize the Markov Decision Process \cite{b17, b18} for optimal classification. Also, we have used only lexical features of the URLs for this experimentation and we would like to explore the performance of our model with the use of other advanced features like host-based features, content-based features etc. or a combination of these various features to build an optimal and robust classifier. There are some other deep learning based algorithms that should be examined for the problem stated in this paper such as LSTM \cite{b15, b16}. Moreover this classifier can be extended for other binary classification problems like Webspam detection \cite{b12} and presence of malicious bots in the network \cite{b13}. RL based approach being more adaptive, the classifier can be extended for mitigating various privacy and security concerns \cite{b14} in wearable devices. 

\section*{Acknowledgement}
\label{sec:acknowledgement}
This project is funded in part by grants (Awards No: 1723765 and 1821560) from National Science Foundation.


\begin{thebibliography}{00}
\bibitem{b4} Abdelhamid, N., Ayesh, A. and Thabtah, F., 2014. Phishing detection based associative classification data mining. Expert Systems with Applications, 41(13), pp.5948-5959.
\bibitem{b12}
Chatterjee, M. and Siami Namin, A., 2018, July. Detecting web spams using evidence theory. IEEE 42nd annual computer software and applications conference (COMPSAC).
\bibitem{b13}
Chatterjee, M., Siami Namin, A. and Datta, P., 2018, December. Evidence Fusion for Malicious Bot Detection in IoT. IEEE International Conference on Big Data (Big Data).
\bibitem{b14}
Datta, P., Siami Namin, A. and Chatterjee, M., 2018, December. A Survey of Privacy Concerns in Wearable Devices. IEEE International Conference on Big Data (Big Data).
\bibitem{b3} Ebbu2017 Phishing Dataset. Accessed 5 April 2019. Available: https://github.com/ebubekirbbr/ pdd/tree/master/input.
\bibitem{b33} Liu, W., Huang., G., Xiaoyue, L. Min, Z., and Deng, X., 2005., Detection of phishing webpages based on visual similarity.  14th international conference on world wide web (WWW).
\bibitem{b34} Nguyen, N. Siami Namin, A., Dang, T. 2018. MalViz: an interactive visualization tool for tracing malware. ISSTA.
\bibitem{b5} Mohammad, R.M., Thabtah, F. and McCluskey, L., 2012, An assessment of features related to phishing websites using an automated technique, IEEE Conference for Internet Technology and Secured Transactions.
\bibitem{b23}
Mnih, V., Kavukcuoglu, K., Silver, D., Rusu, A. A., Veness, J., Bellemare, M. G., Graves, A., Riedmiller, M., Fidjeland, A. K., Ostrovski, G., Petersen, S., Beattie, C., Sadik, A., Antonoglou, I., King, H., Kumaran, D., Wierstra, D., Legg, S., and Hassabis, D. (2015). Human-level control through deep reinforcement learning. Nature, 518(7540):529–533.
\bibitem{b2} Sahingoz, O.K., Buber, E., Demir, O. and Diri, B., 2019. Machine learning based phishing detection from URLs. Expert Systems with Applications, 117, pp.345-357.
\bibitem{b20} Sartoli, S., and Siami Namin, A., 2017, A semantic model for action-based adaptive security, Symposium on Applied Computing (SAC).
\bibitem{b15}
Siami-Namini, S., Tavakoli, N., and Siami Namin, A., 2018, December. A Comparison of {ARIMA} and {LSTM} in Forecasting Time Series. International Conference on Machine Learning and Applications {ICMLA}.
\bibitem{b16}
Siami-Namini, S. and Siami Namin, A., 2018, Forecasting Economics and Financial Time Series: {ARIMA} vs. {LSTM}, CoRR abs/1803.06386.
\bibitem{b11} Sutton, R.S. and Barto, A.G., 2018. Reinforcement learning: An introduction. MIT press.
\bibitem{b19}
Tavakoli, N., Dai, Dong, and Chen Y., 2019, Client-side straggler-aware {I/O} scheduler for object-based parallel file systems,  Parallel Computing.    
\bibitem{b10} WHOIS: Search, Domain Name, Website, and IP Tools. https://who.is
\bibitem{b7} Zhang, Y., Hong, J.I. and Cranor, L.F., 2007, May. Cantina: a content-based approach to detecting phishing web sites. In Proceedings of the ACM conference on World Wide Web.
\bibitem{b17}
Zheng, J. and Siami Namin, A., 2018, A Markov Decision Process to Determine Optimal Policies in Moving Target, Proceedings of the ACM SIGSAC Conference on Computer and Communications Security.
\bibitem{b18} 
Zheng, J. and Siami Namin, A., 2018, Defending {SDN}-based {I}o{T} Networks Against DDoS Attacks Using Markov Decision Process, {IEEE} Conference on Big Data. 
\bibitem{b8} Xiang, G., Hong, J., Rose, C.P. and Cranor, L., 2011. Cantina+: A feature-rich machine learning framework for detecting phishing web sites. ACM Transactions on Information and System Security (TISSEC).








\end{thebibliography}
\end{document}